\documentclass[fleqn,10pt]{wlscirep}
\usepackage{cite}
\usepackage{amsmath,amssymb,amsfonts}
\usepackage{algorithmic}
\usepackage{graphicx}
\usepackage{textcomp}
\usepackage{xcolor}
\usepackage{enumitem}
\usepackage{lipsum}
\usepackage{titlesec}
\usepackage [english]{babel}
\usepackage [autostyle, english = american]{csquotes}
\usepackage{subcaption}
\usepackage[numbers, compress]{natbib}
\usepackage{array}
\usepackage{tikz}

\MakeOuterQuote{"}

\titlespacing*{\section}{0pt}{1 ex plus .5 ex minus .2ex}{1 ex plus .2ex}
\titlespacing*{\subsection}{0pt}{1 ex plus .5 ex minus .2ex}{1 ex plus .2ex}
\titlespacing*{\subsubsection}{0pt}{1 ex plus .5 ex minus .2ex}{1 ex plus .2ex}

\def\BibTeX{{\rm B\kern-.05em{\sc i\kern-.025em b}\kern-.08em
    T\kern-.1667em\lower.7ex\hbox{E}\kern-.125emX}}

\begin{document}

\title{Leveraging Self-Supervised Features for Efficient Flooded Region Identification in UAV Aerial Images}



\author[1]{Dibyabha Deb}
\author[2,*]{Ujjwal Verma}
\affil[1]{Department of Electronics and Communication Engineering, Manipal Institute of Technology Bengaluru, Manipal Academy of Higher Education, Manipal, India}
\affil[2]{Department of Electronics and Communication Engineering, Manipal Institute of Technology, Manipal Academy of Higher Education, Manipal, India}
\affil[*]{Corresponding Author: ujjwal.verma@manipal.edu}

\begin{abstract}
Identifying regions affected by disasters is a vital step in effectively managing and planning relief and rescue efforts. Unlike the traditional approaches of manually assessing post-disaster damage, analyzing images of Unmanned Aerial Vehicles (UAVs) offers an objective and reliable way to assess the damage. In the past, segmentation techniques have been adopted to identify post-flood damage in UAV aerial images. However, most of these supervised learning approaches rely on manually annotated datasets. Indeed, annotating images is a time-consuming and error-prone task that requires domain expertise. This work focuses on leveraging self-supervised features to accurately identify flooded regions in UAV aerial images. This work proposes two encoder-decoder-based segmentation approaches, which integrate the visual features learned from DINOv2 with the traditional encoder backbone. This study investigates the generalization of self-supervised features for UAV aerial images. Specifically, we evaluate the effectiveness of features from the DINOv2 model, trained on non-aerial images, for segmenting aerial images, noting the distinct perspectives between the two image types. Our results demonstrate that DINOv2's self-supervised pretraining on natural images generates transferable, general-purpose visual features that streamline the development of aerial segmentation workflows. By leveraging these features as a foundation, we significantly reduce reliance on labor-intensive manual annotation processes, enabling high-accuracy segmentation with limited labeled aerial data.

The code utilized for the analyzes mentioned in this paper is publicly available on GitHub. To access the repository, visit the link (\href{https://github.com/Dibyabha/uav-floodnet-ss}{GitHub Link}). The repository includes detailed descriptions and instructions for replicating the results presented in the paper.
\end{abstract}
\flushbottom
\maketitle

\section{Introduction}

Floods are among the most destructive natural disasters, causing significant economic damage, environmental impacts, and loss of human life. They constitute a substantial part of the climate change-induced stressors on water resources globally. In recent years, the frequency and severity of flood events have increased due to various man-made and natural factors, making flood risk management a critical area of research. This group of natural disasters has been one of the leading causes of death and economic losses throughout the globe. Studies in this field have shown a significant rise in the death toll as well as financial losses with 399 total recorded natural disasters, which resulted in 86,473 fatalities and affected 93.1 million people, amounting to the economic losses of US{\$}202.7 billion \cite{2023_EMDAT_report}. Previous studies have shown that around 40 million hectares of land in India are subjected to flood-prone, with an annual average of 18.6 million hectares of land affected \cite{flooddisman}.

Identifying the flooded region is essential in post-disaster search and rescue efforts. Traditional methods to assess the damage typically include surveying and manually visiting the flood-affected areas. Many studies rely on traditional statistical methods, which may not fully capture the complex dynamics of flood events \cite{limit}. Conventional techniques are also influenced by human bias and human error, and it takes a lot of time to complete the surveying process. These challenges led researchers to analyze statistical measures along with remote sensing for assessing the flood damage \cite{reviewflood}.

Data acquired from remote sensing techniques such as Earth Observation (EO) satellites and Unmanned Aerial Vehicles (UAVs) provide an alternative approach to monitor and estimate post-disaster damage. EO, through the use of satellites, provides a global view that is particularly useful for monitoring large-scale environmental phenomena. It allows for the continuous collection of data over both time and space, enabling researchers to track the development and progression of flood events. This can lead to better predictions and, ultimately, more effective flood management strategies by incorporating satellite data \cite{uavsat1}. Moreover, the data obtained from EO are objective and reliable and are free from biases and errors, unlike traditional survey methods. Along with that, EO also offers a safe and efficient way to gather data even from remote or hard-to-reach areas. Due to their ability to rapidly deploy and customize flight paths, UAVs offer a more localized view and can complement the satellite-based images for disaster assessment \cite{dronetech1}.  

In the past, semantic segmentation approaches have been utilized to assess the damage caused by flood from UAV images \cite{uavsat1} \cite{uav2}. However, these approaches focused on supervised segmentation approaches, which require a huge amount of manually annotated data for training these models. Indeed, annotating images is a time-consuming and error-prone task that requires domain expertise. Recently, self-supervised learning has shown promising results in the segmentation of images for various applications such as autonomous vehicles \cite{ozbulak2023knowtmlrreview} \cite{PAMIReview2021}. Unlike supervised learning, self-supervised learning does not require a large amount of training data. For instance, DINOv2 can learn rich general-purpose visual features by pretraining on a large quantity of curated data and can be used for multiple downstream tasks without the need for fine-tuning \cite{oquab2024dinov}. The limitations associated with supervised learning can be overcome by using such self-supervised learning techniques. The major advantage of using such methods is that they generate high-dimensional feature vectors from the images without fine-tuning, and the pre-trained weights can be utilized for feature extraction in images that are not similar to the images on which these pre-trained models are trained. Utilizing this property of transferring the features from a self-supervised method and combining it with the features extracted using an encoder-decoder architecture can yield better results even on datasets on which the self-supervised methods are not trained.

In this work, we propose an encoder-decoder-based semantic segmentation architecture for flood region identification in UAV aerial images.  The proposed architecture contains an encoder module that fuses the visual features learned without supervision with the standard segmentation encoder module. Subsequently, the fused feature is fed to the decoder module to generate the semantic segmentation map of the flooded region. Specifically, this work proposes two architectures: The first architecture integrates the DINOv2 features in the U-Net segmentation model, while the second architecture integrates DINOv2 features in the DeepLabV3 model. This study investigates the generalization of self-supervised features for UAV aerial images. Specifically, we evaluate the effectiveness of features from the DINOv2 model, trained on non-aerial images, for segmenting aerial images, noting the distinct perspectives between the two image types. This study addresses the critical question:\textit{ To what extent do self-supervised features learned on non-aerial images generalize to aerial image domains? }This investigation is motivated by the fundamental shift in visual perspective between aerial imagery (characterized by top-down orthogonal views) and natural images (typically captured from oblique or eye-level viewpoints).

To the best of our knowledge, this is the first time such a combination has been employed for flood detection. Such integration of a self-supervised method with an encoder-decoder architecture significantly reduces the computational resources and time required for training while maintaining high accuracy. This work also proposes a weighted focal loss function. This weighted focal loss incorporates inverse class frequency to address the class imbalance. This class-frequency-based weighting ensures that underrepresented classes receive greater emphasis during training, thereby improving segmentation accuracy for minority classes such as flooded regions. 

 

This article is organized as follows. Section II summarizes the recent works in the identification of flood-affected regions using semantic segmentation of UAV aerial images. Section III describes the proposed architecture in detail. Section IV and V presents the experiments and various results obtained.


\section{Related Work}

The study proposes an encoder-decoder-based semantic segmentation architecture for the identification of flooded regions in UAV aerial images utilizing a pre trained self-supervised method. This section briefly discusses the recent advances in semantic segmentation of UAV aerial images and post-disaster scene understanding.

\subsection{Semantic Segmentation}

Semantic segmentation is the process of assigning different class labels to individual pixels in an image. This can be applied to the images collected by UAVs and EO satellites to identify and classify different elements of the image, such as water, vegetation, roads, and buildings, by the classification of each pixel into one of many categories. Semantic segmentation is also widely used in other applications such as medical image analysis \cite{unet} \cite{artcnn}, geographic information systems, autonomous driving \cite{gis} \cite{auto}, etc. 

Deep learning techniques such as Convolutional Neural Networks (CNNs) and Recurrent Neural Networks (RNNs), as well as statistical methods such as regression analysis and time series analysis, have been extensively employed for flood-related analysis using aerial video data. Specifically, CNNs demonstrate strong capabilities in processing drone imagery to detect hydrological features such as water extent and inundated regions \cite{hess-26-4345-2022} \cite{8517946}, while RNNs excel in temporal pattern recognition for flood event forecasting through historical data sequences. Complementary to these, statistical techniques like regression analysis provide quantitative estimation of hydrological parameters including water depth and flood-affected areas, whereas time series analysis enables probabilistic forecasting of future flood occurrences through spatiotemporal pattern analysis of historical events\cite{regflood}.


In the last decade, deep learning-based semantic segmentation approaches have shown remarkable success and outperformed the traditional segmentation approaches. Fully Convolutional Network (FCN), a neural network that only performs convolution (and subsampling or upsampling) operations, is the foundation of the deep learning-based semantic segmentation task, which has been applied for various applications \cite{fcn}. SegNet follows the model of VGG16 and categorizes low-level pixel values to high-order semantic information \cite{badrinarayanan2017segnet}. U-Net has significantly improved the feature extraction by stitching together the output information from the encoder layer to the corresponding decoder layer during upsampling \cite{unet} \cite{semsegunet}. DeepLab proposed and developed an atrous spatial pyramid pooling (ASPP) module to obtain multiscale information meanwhile maintaining high-resolution feature maps \cite{deeplab} \cite{deeplabv3}. The RNN models use global information and improve segmentation outputs by associating pixel-level information with local information \cite{rnn}. However, these supervised learning-based models require manually annotated data for training. Indeed, annotating each individual pixel in an image is a time-consuming and error-prone task. Besides, domain knowledge is required for annotating images for certain applications like Earth Observation Images, and Medical Images. 

In the last few years, transformers have completely revolutionized the domain of Natural Language Processing (NLP). With its success in NLP, researchers have begun exploring the objective of using transformers in the domain of computer vision. Vision Transformer (ViT) was one of the first transformers developed with this objective which uses a fully transformer (deep learning model that utilizes the self-attention mechanism by differentially weighting each part of the data) based design for classification and other image recognition tasks \cite{dosovitskiy2021an}. The introduction of ViT led to a dramatic increase in the performance of computer vision tasks which also led researchers to work on the improvement of accuracy. Data-efficient image Transformer (DeiT) was introduced with this objective which has a teacher-student framework \cite{pmlr-v139-touvron21a}. After its introduction, many architectures were designed to solve dense prediction tasks such as image segmentation and object detection. Segmentation Transformer (SETR) used ViT as an encoder and developed different variants of decoder designs \cite{rethink}. On top of a ViT-based encoder, Segmenter designed a mask-based decoder network \cite{segmenter}. However, the transformer-based approach requires a large amount of manually annotated data for training and generalizes poorly for small datasets.

Recently, self-supervised approaches have shown promising results, especially by reducing the dependence on the availability of manually annotated data. Self-supervised learning (SSL) allows models to learn from vast amounts of unlabeled data, which is both abundant and inexpensive compared to labeled data. This capability is crucial because labeled datasets require manual annotation which is labor-intensive and costly. SSL enables the extraction of more generalizable and robust features which are learned by solving pretext tasks designed to predict parts of the data from other parts, capturing intrinsic structures and patterns in the data. As new data becomes available, self-supervised models can be updated and refined without the need for additional labeled data, making them highly adaptable to changing environments and new tasks. In \cite{wang2022fully}, the authors developed a fully self-supervised model for image semantic segmentation using a bootstrapped training scheme that leverages global semantic knowledge for self-supervision. In \cite{oquab2024dinov}, the authors developed a self-supervised method that is designed to learn robust visual features without supervision. The method involves a discriminative self-supervised pretraining approach and is designed to stabilize and accelerate training, especially when scaling to larger models and datasets. The pipeline used to curate the dataset uses data similarities, eliminating the need for manual annotation, and models are based on a slightly modified version of the ViT model, with different sizes and architectures optimized for various applications.

Along with the utilization of deep learning-based methods, transfer learning is often used to address the issue of data scarcity and to improve the target task performance with less training time by using the pre-trained network. Transfer learning uses a pre-trained network as a starting point to learn a new task where the method itself does not include training but does need to be fine-tuned to get reasonable accuracy and precision.

Semantic Segmentation models have also been utilized for analyzing UAV aerial images \cite{CHENG20241}. UNetFormer has a ResNet-based encoder, and the decoder is based on a transformer model \cite{unetlike}. The authors developed a global-local attention mechanism to model both global and local information in the decoder phase. Road Structure Refined CNN (RSRCNN) model was developed for extracting road structure features from aerial scenes \cite{road} where the primary purpose was to create a specific segmentation model to extract road structures along with defining a new loss function for the particular task. A Deep CNN (DCNN) model was proposed for semantic segmentation to eliminate or counter the blurry object boundary by combining semantic segmentation with semantically informed edge detection \cite{dcnn}. Attentive Bilateral Contextual Network (ABCNet) was proposed and developed to use a dual path approach to capture long-range and fine-grained information \cite{LI202184}. In \cite{LI202184}, the authors developed a dual pathway where one branch was used to retain affluent spatial details and the other branch to capture global contextual information. DINOv2-based models have been previously used for self-localization in low attitude environments \cite{dino-low}. Prompt-based methods for flood detection and segmentation from UAV images have also been explored in previous studies \cite{prompt}.

\subsection{Natural Disaster Damage Assessment}

Researchers have also studied and estimated the future flood damage quantification analysis for building structures \cite{futflood}, and the analysis and quantification of future flood damage and influencing factors in urban areas \cite{quantflood}. But, oftentimes, such methods involve the use of traditional surveying or statistical methods, which are limited by the global terrain that aggravates the problem of sampling distribution. These methods employ many workers to collect data points from a site, leading to an increase in the per-person payment and the total cost of the project \cite{dronetech1}. These traditional methods rely on a non-probabilistic approach and, hence, cannot yield a valid estimate of the risks of error. Determining bias using this procedure is also opposed by the fact that there is a general lack of explicit description of the sampling methodology in traditional surveys \cite{limit}. Conventional methods are also influenced by human bias and human error, and it takes a long time to complete the surveying process.


To address the challenges, UAV and EO images have been used to detect the presence of man-made features and, hence, understand the region of maximum impact in the occurrence of a natural disaster \cite{doshi2018satellite}. Recent works have also attempted to estimate the post-disaster damage by analyzing UAV Aerial images \cite{VermaSurveyFlood} \cite{AnanyaFlood2021} \cite{8517946}. For instance, an unsupervised learning approach is used to estimate the severity of flooding \cite{AnanyaFlood2021}. Multilevel Instance Segmentation Network (MSNet) was also proposed to assess aerial videos for building damage after any natural disaster \cite{msnet}. Modified U-Net architecture with attention mechanism was also proposed and developed for disaster management and assessment \cite{unetattn}. Unlike existing methods that require a large number of manually annotated images, the proposed approach integrates self-supervised features into the traditional encoder-decoder-based semantic segmentation model, thereby reducing the dependence on manually annotated images. Furthermore, the proposed work examines the efficacy of self-supervised features derived from non-aerial images to segment aerial images, taking into account the distinct perspectives of the two image types for post-disaster scene understanding.


\begin{figure}
    \centering
    \includegraphics[width=0.9\linewidth]{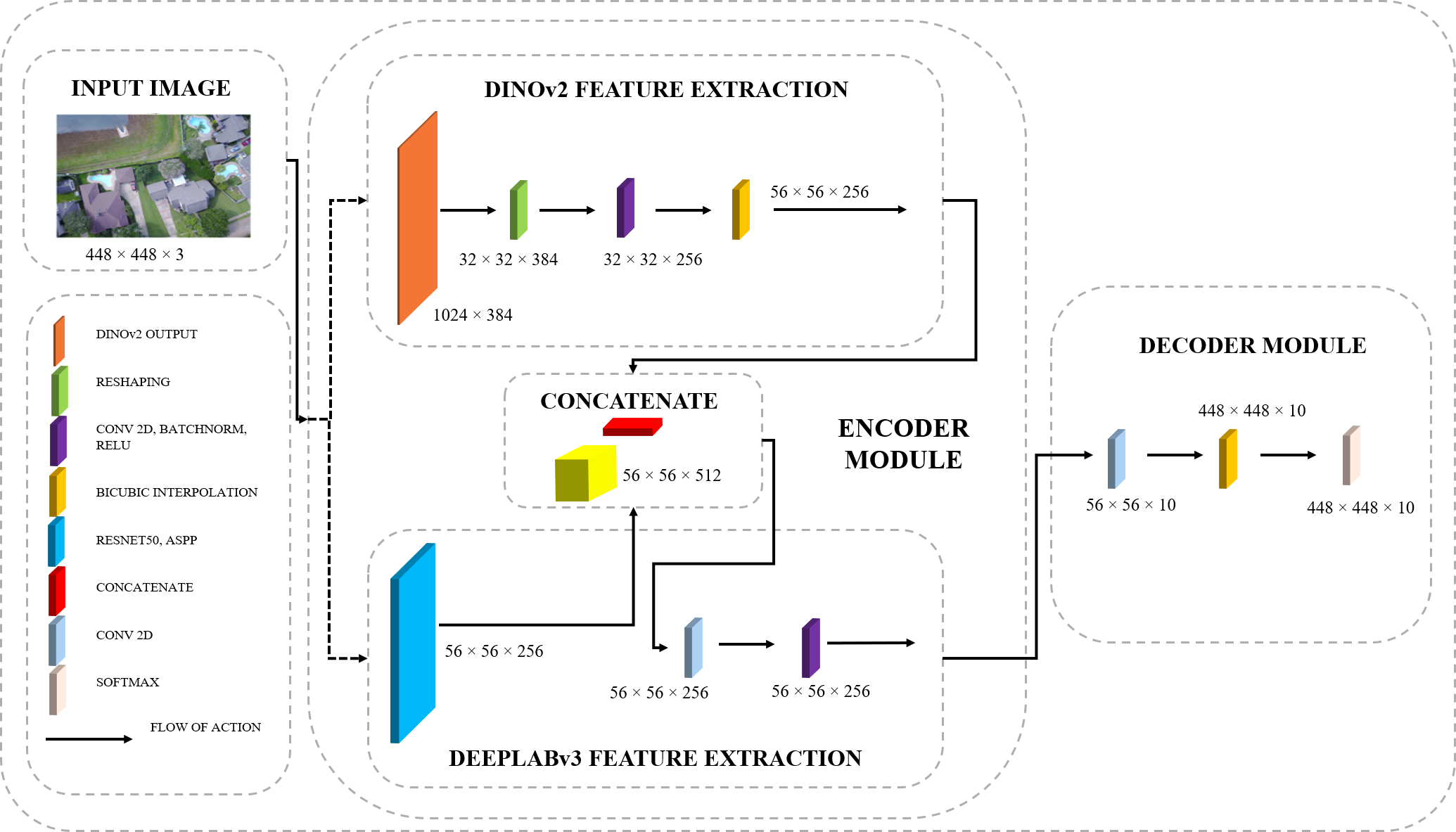}
    \caption{Integrating Self-Supervised Features in DeepLab: The features from DINOv2 module is integrated into the DeepLab encoder and then processed through a decoder module to generate the final segmentation map. }
    \label{fig:DeepLabDinoMethod}
\end{figure}

\begin{figure}
    \centering
    \includegraphics[width=0.9\linewidth]{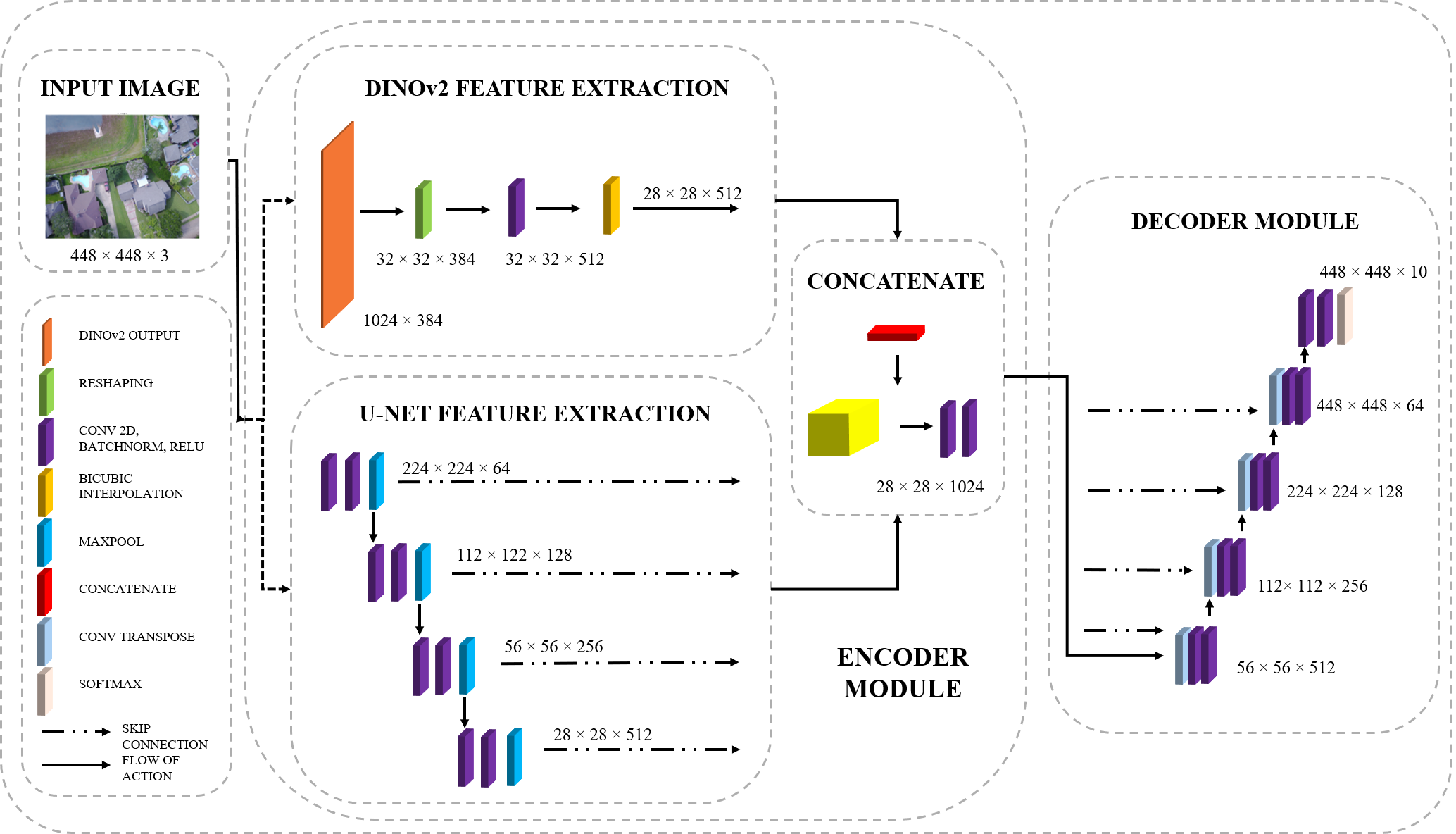}
    \caption{Integrating Self-Supervised Features in U-Net: The features from DINOv2 module and U-Net encoder are fused together and then fed to the decoder module to create the final segmentation.  } 
    \label{fig:UNetDinoMethod}
\end{figure}

\section{Methodology}
This section discusses in detail the proposed approach for the semantic segmentation of high-resolution UAV images. The present work proposes two encoder-decoder architectures, which are obtained by fusing the DINOv2 pre-trained features with U-Net \cite{unet} and DeepLabV3 \cite{deeplabv3}. The primary goal is to evaluate the effectiveness of self-supervised features (DINOv2) as a feature extractor for downstream computer vision tasks, particularly UAV image segmentation, for post-disaster scene understanding. Importantly, DINOv2 is used without fine-tuning, relying solely on its pre-trained weights to extract robust visual features. This strategy aims to evaluate the model's generalization capability when applied to high-resolution UAV images, which differ significantly from its training datasets, which are natural images. The overview of our proposed approaches is illustrated in Figures \ref{fig:DeepLabDinoMethod} and  \ref{fig:UNetDinoMethod}.

The overview of the proposed approach for the segmentation of UAV images is as follows:  The input image was simultaneously fed into two distinct modules: a DINOv2 pre-trained model and an encoder module. The features extracted by both modules were then concatenated in the encoder bottleneck, creating a unified and enriched feature set. This concatenated feature vector was subsequently processed through the decoder module, which generated the predicted segmentation map. The decoder's role is to interpret the combined features and produce a detailed segmentation output that accurately delineates various regions or objects within the UAV image. 

The details of the two encoder-decoder architectures, viz., DINOv2-U-Net and DINOv2-DeepLabV3, are discussed below.

\subsection{DINOv2-U-Net Feature Extraction}

The DINOv2 pre-trained model was employed for feature extraction due to its ability to learn representations from unlabeled data, eliminating the need for manual annotations. During the training phase, the pre-trained model's parameters were frozen, ensuring it did not fine-tune for the specific task. Instead, feature extraction relied solely on the model's previously learned weights. The pre-processed image, $I$, was divided into smaller patches (each image was subdivided into $P$ patches) by the pre-trained model. The model then processed each of the patches to generate patch embeddings. The normalized features were returned, which were reshaped. Fundamentally, the function of the pretrained model was to take the input image and create $P$ non-overlapping patches of the image each having dimension of $p \times p$. The output from the pre-trained model consists of patch embeddings of all the patches of all the images in the batch. The final dimension of the high dimensional feature vector, after reshaping, becomes $B \times H/p \times W/p \times E$, where $B$ represents the total number of images in a batch, $E$ referred to the patch embeddings of each patch of each image in the batch and $H/p \times W/p$ referred to the total number of patches $P$. The features extracted were convolved with a kernel size of $1 \times 1$ and a stride of $1$ to match the dimensionality of the U-Net encoder bottleneck.

The U-Net encoder took the pre-processed input image $I$ and compressed the spatial dimensions while enhancing feature depth by increasing the feature channels, typically doubling the features starting from 64, which progressively captured more abstract and high-level features of the image. The U-Net encoder was represented by a series of convolutions and subsequent pooling operations. During this process, there were two consecutive convolution layers (3$\times$3) with batch normalization and ReLU activation function. The activation in each layer was passed through a max-pooling layer (2$\times$2), which reduced the spatial dimension to extract the essential information and pass it along to the next layer in the encoder. At the bottleneck of the U-Net, a convolution operation was performed on 512 feature maps, connecting the encoder to the decoder. Before the convolution at the bottleneck, the DINOv2 features and the U-Net features were concatenated.

The U-Net decoder was the counterpart to the U-Net encoder, which was responsible for reconstructing the spatial dimensions back to the original size of the image. It up-sampled (through transposed convolutions) the concatenated feature maps of DINOv2 and U-Net encoder and linked the corresponding layers from the encoder to the decoder using skip connections which helped to recover the spatial resolution and finer details that might be lost during the down-sampling process in the encoder. The final output from the decoder, after applying softmax, was the predicted segmentation mask. This mask was compared with the ground truth mask using a custom loss function. The overview of the method is shown above in Figure \ref{fig:UNetDinoMethod}.

\subsection{DINOv2-DeepLabV3 Feature Extraction}


Similar to the DINOv2-U-Net Feature Extraction discussed above, the pretrained DINOv2 model was utilized to learn features without manually annotated data. The DeepLabV3 encoder, which took the pre-processed input image $I$, leveraged a pretrained ResNet-50 model. The last layer of the encoder was adjusted to match the total number of classes in the dataset, rather than the 21 classes it was originally trained on. This modification ensured that the encoder extracted features tailored to the specific segmentation requirements of the dataset. The features extracted after the ASPP module were concatenated with the DINOv2 features. The concatenated features were further convolved and then applied to the decoder. At the bottleneck, a convolution operation was performed on 256 feature maps, which were then concatenated with the DeepLabV3 feature maps. 

The DeepLabV3 decoder up-sampled the feature maps directly from the last layer of the encoder stage. The up-sampled feature map was sent through a softmax layer and was subsequently compared with the ground truth mask using a custom loss function. The overview of the method is shown above in Figure \ref{fig:DeepLabDinoMethod}

\section{Experiments}
\subsection{Image Dataset}

The proposed approach was evaluated on the FloodNet Dataset \cite{dataset}, which contained high-resolution Unmanned Aircraft Systems (UAS) imagery with detailed semantic annotation of flooded and non-flooded regions. Researchers collected the data after Hurricane Harvey using a small UAS platform, DJI Mavic Pro quadcopters. All images demonstrated post-flooded damages to affected areas. The dataset consisted of 2343 images with dimensions of 3000 $\times$ 4000 $\times$ 3, divided into training sets (\(\sim\)60{\%}, 1445 images), validation sets (\(\sim\)20{\%}, 450 images), and test sets (\(\sim\)20{\%}, 448 images). The dataset included 10 different classes: 'Background,' 'Building Flooded,' 'Building Non-Flooded,' 'Road Flooded,' 'Road Non-Flooded,' 'Water,' 'Tree,' 'Vehicle,' 'Pool,' and 'Grass.'

\subsection{Pre-Processing}
The images used for our work were represented as $I$, where \textit{I} stood for an input image. Under the image pre-processing step, we converted all the images, $I$, to have the same height and width. Essentially, all the images, along with their ground truth masks, were pre-processed to a particular dimension of \(\textit{H} \times \textit{W} \times \textit{c}\), where $H$ and $W$ were the height and width of the image, respectively and \textit{c} was the number of channels (3) in the image. Along with adjusting the height and width, we normalized the images to a scale of 0 to 1 and applied augmentation strategies such as random horizontal flip, random vertical flip, and zooming to overcome the over-fitting problem during the training phase. We resized the input images to dimensions of $448 \times 448$ pixels using bicubic interpolation. 

\subsection{Loss Function and Hyper-Parameters}
In this study, a custom loss function is proposed to enhance the performance of the segmentation model. Specifically, a weighted focal loss is introduced, which incorporates inverse class frequency to address the class imbalance in the traditional focal loss \cite{focalloss}. It is observed that the number of images in the different classes of FloodNet dataset is not uniform and suffers from class imbalance (Table \ref{tab:alpha_values}). Traditional losses fail here, but our weighted focal loss reduces bias by penalizing misclassifications in sparse classes more heavily. To mitigate this imbalance, the proposed method assigns weights to the focal loss based on the number of images in each class. This class-frequency-based weighting ensures that underrepresented classes receive greater emphasis during training, thereby improving segmentation accuracy for minority classes. The proposed weighted focal loss is combined with the dice loss \cite{diceloss} to ensure that the model focuses on rare classes (e.g., flood-damaged structures) without overfitting to dominant ones (e.g., water bodies).


\begin{table}[h!]
    \centering
    \begin{tabular}{|c|c|c|}
        \hline
        \textbf{Class} &\textbf{Class Name} &\textbf{Alpha} \\ \hline
        Class 0 & Background & 0.08 \\ \hline
        Class 1 & Building Flooded & 0.14 \\ \hline
        Class 2 & Building Non-Flooded & 0.10 \\ \hline
        Class 3 & Road Flooded & 0.10 \\ \hline
        Class 4 & Road Non-Flooded & 0.05 \\ \hline
        Class 5 & Water & 0.015 \\ \hline
        Class 6 & Tree & 0.01 \\ \hline
        Class 7 & Vehicle & 0.25 \\ \hline
        Class 8 & Pool & 0.25 \\ \hline
        Class 9 & Grass & 0.005 \\ \hline
    \end{tabular}
    \caption{Alpha ($\alpha$) value per class for class-frequency weighted focal loss.}
    \label{tab:alpha_values}
\end{table}

The proposed weighted focal loss is incorporated to address class imbalance by down-weighting easy-to-classify examples. The loss is defined as:

\begin{equation}
\text{Focal Loss}_{\alpha} = -\alpha \cdot y_{\text{true}} \cdot \log(y_{\text{pred}})\cdot (1 - y_{\text{pred}})^{\gamma},
\end{equation} where, \( \alpha \) is a balancing factor for each class, \( \gamma \) is a focusing parameter that adjusted the rate at which each examples were down-weighted and \(y_{true}\) and \(y_{pred}\) are the ground truth and predicted masks, respectively.

Alpha ($\alpha$) values were derived based on the inverse frequency of each class in the training set. This ensured that rarer classes receive higher weights, counteracting their under-representation during training. The alpha values were calculated using inverse class frequencies from the training dataset. The inverse frequency for each class \( i \) is computed as:

\begin{equation}
\text{inv\_freq}_i = \frac{1}{\text{count}_i + 10^{-6}},
\end{equation}

where \( \text{count}_i \) represented the number of pixels belonging to class \( i \), and a small constant \( 10^{-6} \) was added to avoid division by zero. The inverse frequencies were normalized to ensure their sum equaled one, resulting in the final alpha values:

\begin{equation}
\alpha_i = \frac{\text{inv\_freq}_i}{\sum_{j} \text{inv\_freq}_j} + \beta,
\end{equation}

In the above equation, the value of $\beta$ was selected between -0.20 and 0.20 for each class separately. The final alpha values are reported in a tabular format in Table \ref{tab:alpha_values}.

The dice loss \cite{diceloss} is designed to measure the overlap between the predicted segmentation and the ground truth. It is particularly effective for imbalanced datasets, as it emphasized the importance of correctly predicting smaller classes. The Dice loss is computed as follows:

\begin{equation}
\text{Dice Loss} = 1 - \frac{2 \cdot {inter} + {\epsilon}}{{union} + {\epsilon}},
\end{equation} where, \( inter \) is the intersection of the predicted \(y_{pred}\)  and true masks \(y_{true}\), \( union \) is the sum of the predicted \(y_{pred}\) and true masks \(y_{true}\), ( $\epsilon$) is a small constant added to prevent division by zero.

The total loss function combined both dice loss and weighted focal loss, weighted by a hyperparameter to balance their contributions during training:

\begin{equation}
\text{Total Loss} = w \cdot \text{Dice Loss} + (1 - w) \cdot \text{Focal Loss}_{\alpha}, 
\end{equation} where \( w \) was a hyperparameter that determined the relative importance of each component. By adjusting this weight, we fine-tuned the model's sensitivity to different aspects of the segmentation task. The hyperparameter \(w\) was set at 0.5 and along with that, \(\gamma\) was set at 4.0 and \(\epsilon\) at $10^{-5}$. 

In our study, we have used the DINOv2 pretrained model using ViT-S14 backbone, which sets the value of \textit{p} to 14. The learning rate was initially kept at $10^{-3}$ during the training phase. A learning rate reduction strategy was implemented, which reduced the learning rate by a factor of 0.75 if the validation loss plateaued for patience of 5 epochs. Adam optimizer was employed to optimize the model parameters. During training, a batch size, \textit{B}, of 8 was used, while a batch size, \textit{B}, of 16 was employed for validation and testing. All models were evaluated over a total of 70 epochs. The \textit{H} and \textit{W} was transformed to 448 and 448, respectively. The total number of patch embeddings, \textit{E}, were 384. Additionally, bicubic interpolation was utilized for all up-sampling operations within the models.

\begin{figure*}[h!]
    \centering
    \begin{tikzpicture}
        \node[anchor=south west, inner sep=0] (image) at (0,0) {\includegraphics[width=\textwidth]{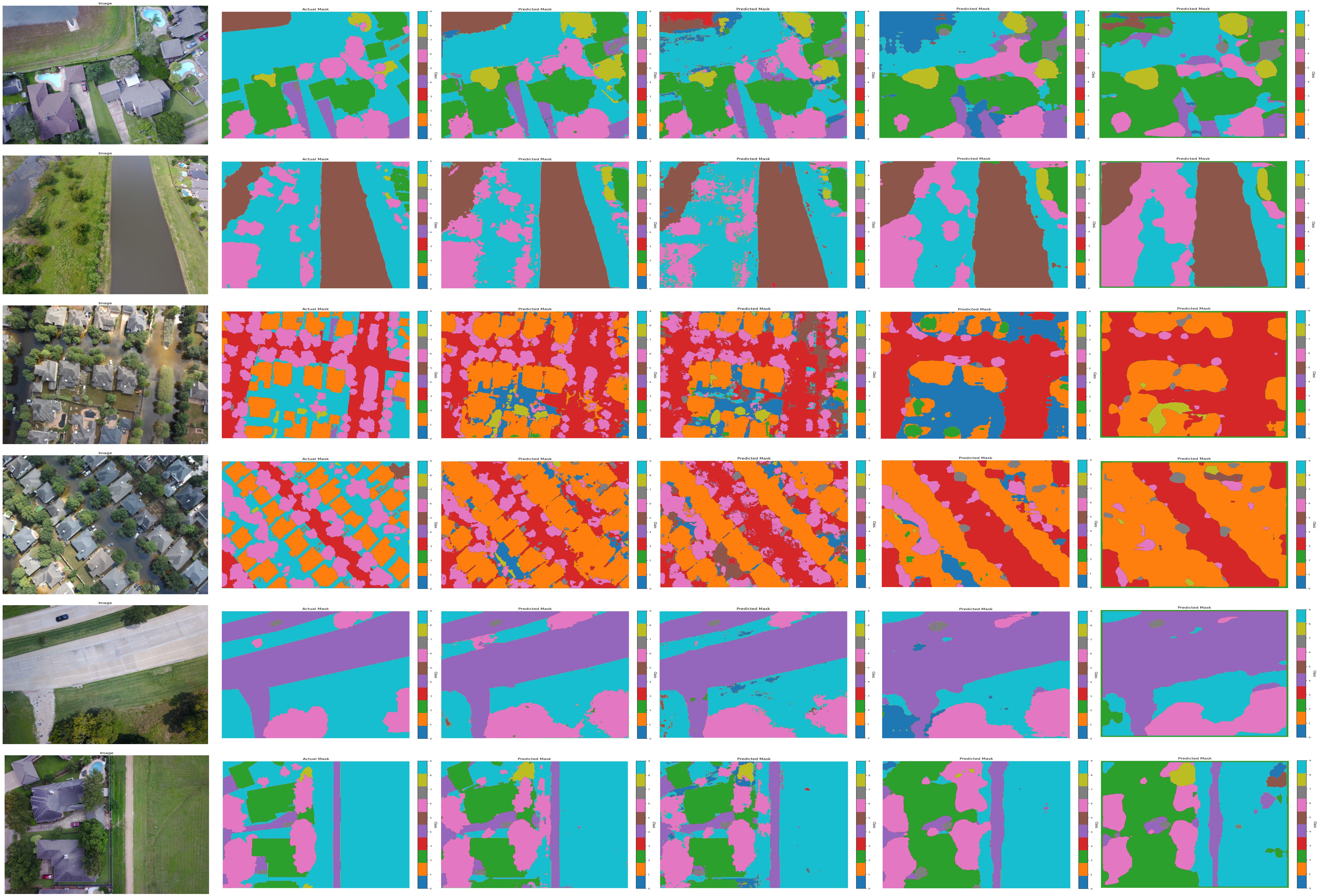}};
        \begin{scope}[x={(image.south east)},y={(image.north west)}]
            \node at (0.08, -0.02) {(a)};
            \node at (0.25, -0.02) {(b)};
            \node at (0.42, -0.02) {(c)};
            \node at (0.58, -0.02) {(d)};
            \node at (0.75, -0.02) {(e)};
            \node at (0.92, -0.02) {(f)};
        \end{scope}
    \end{tikzpicture}

    \caption{Qualitative comparison of the predicted masks across various models. (a) Input image to different models and (b) corresponding ground truth images. (c) Results from DINOv2 with U-Net model, (d) Results from Vanilla U-Net (e), Results from DINOv2 with DeepLabV3 model and (f) Results from DeepLabV3.}
    \label{fig:qualitative_comparison}
\end{figure*}
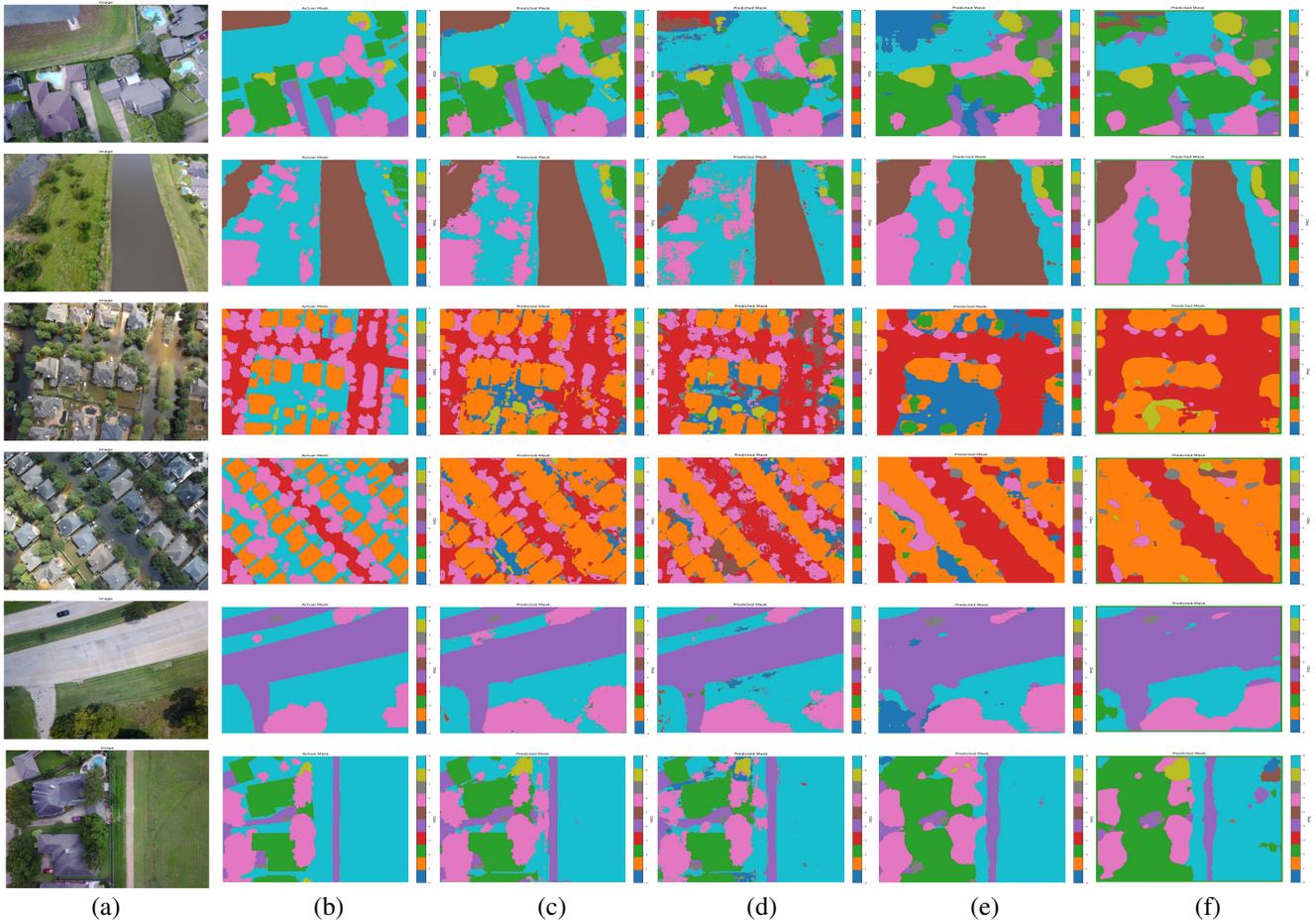

\section{Results and Discussion}

\subsection{Quantitative Analysis}

The proposed approach was compared with existing encoder-decoder architectures in semantic segmentation approaches such as Vanilla U-Net \cite{unet}, DeepLabV3 \cite{deeplabv3}, and E-Net \cite{enet} \cite{dataset}. The mean Intersection over Union (mIoU) metric and class-wise Intersection over Union (IoU) are utilized to compare the performance of the proposed approach with existing methods. The results of our findings are presented in a tabular format in Table \ref{tab:miou_comparison}.

\begin{table*}[htbp]
    \centering
    \begin{tabular}{|>{\centering\arraybackslash}p{2.5cm}|>{\centering\arraybackslash}p{1.2cm}|>{\centering\arraybackslash}p{1.2cm}|>{\centering\arraybackslash}p{1.2cm}|>{\centering\arraybackslash}p{1.2cm}|>{\centering\arraybackslash}p{1.2cm}|>{\centering\arraybackslash}p{1.2cm}|>{\centering\arraybackslash}p{1.2cm}|>{\centering\arraybackslash}p{1.2cm}|>{\centering\arraybackslash}p{1.2cm}|>{\centering\arraybackslash}p{1.2cm}|}
    \hline
        \textbf{Method} & \textbf{Building Flooded} & \textbf{Building Non-Flooded} & \textbf{Road Flooded} & \textbf{Road Non-Flooded} & \textbf{Water} & \textbf{Tree} & \textbf{Vehicle} & \textbf{Pool} & \textbf{Grass} & \textbf{mIoU} \\ 
    \hline
        \textbf{E-Net} & 21.82 & 41.41 & 14.76 & 52.53 & 47.14 & 62.56 & 26.21 & 16.57 & 75.57 & 39.84 \\ 
    \hline
        \textbf{U-Net} & 41.00 & 66.48 & 32.55 & 67.54 & 57.30 & 64.10 & 36.65 & 40.38 & 75.23 & 53.47 \\ 
    \hline
        \textbf{DeepLabV3} & 31.00 & 31.84 & 27.25 & 56.65 & 57.98 & 57.85 & 13.58 & 24.22 & 68.93 & 41.03 \\ 
    \hline
        \textbf{DINOv2+U-Net} & 37.42 & 69.92 & 33.78 & 67.87 & 65.98 & 66.63 & 36.97 & 32.70 & 77.06 & 54.26 \\ 
    \hline
        \textbf{DINOv2+} & 33.33 & 55.99 & 33.92 & 58.36 & 65.32 & 63.56 & 16.13 & 28.42 & 71.43 & 47.39 \\ 
        \textbf{DeepLabV3}  & & & & & & & & & & \\
    \hline
    \end{tabular}
    \caption{Performance evaluation of the proposed approach with existing methods in terms of per-class intersection over union (in \%). The mIoU value (in \%) is shown in the last column for different models.  }
    \label{tab:miou_comparison}
\end{table*}


\begin{figure}[h!]
    \centering
    \begin{subfigure}[b]{0.45\textwidth}
        \centering
        \includegraphics[width=\textwidth]{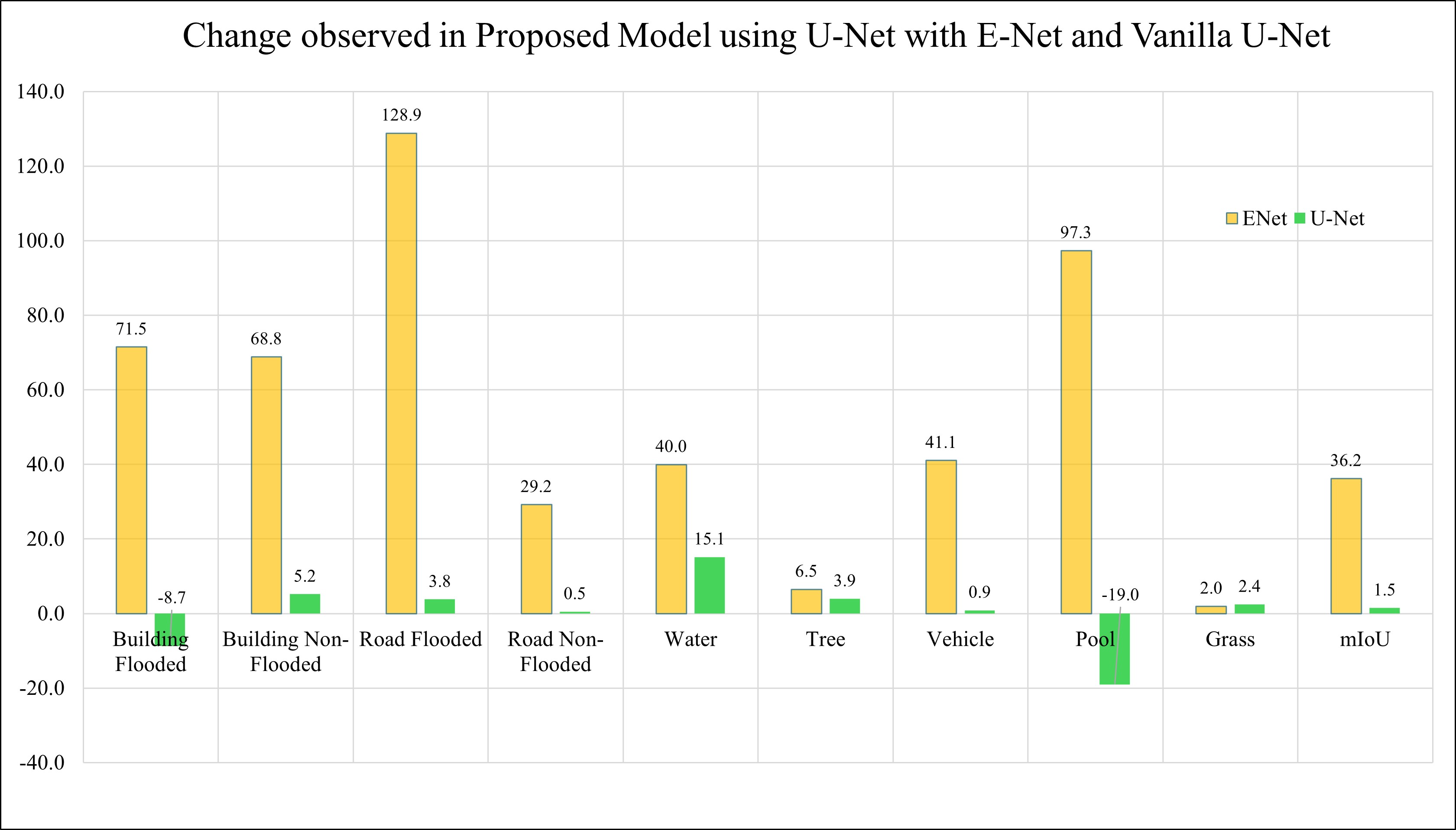}
        \caption{Relative changes observed between proposed model and Vanilla U-Net/E-Net}
        \label{fig:subfig_a}
    \end{subfigure}
    \hfill
    \begin{subfigure}[b]{0.45\textwidth}
        \centering
        \includegraphics[width=\textwidth]{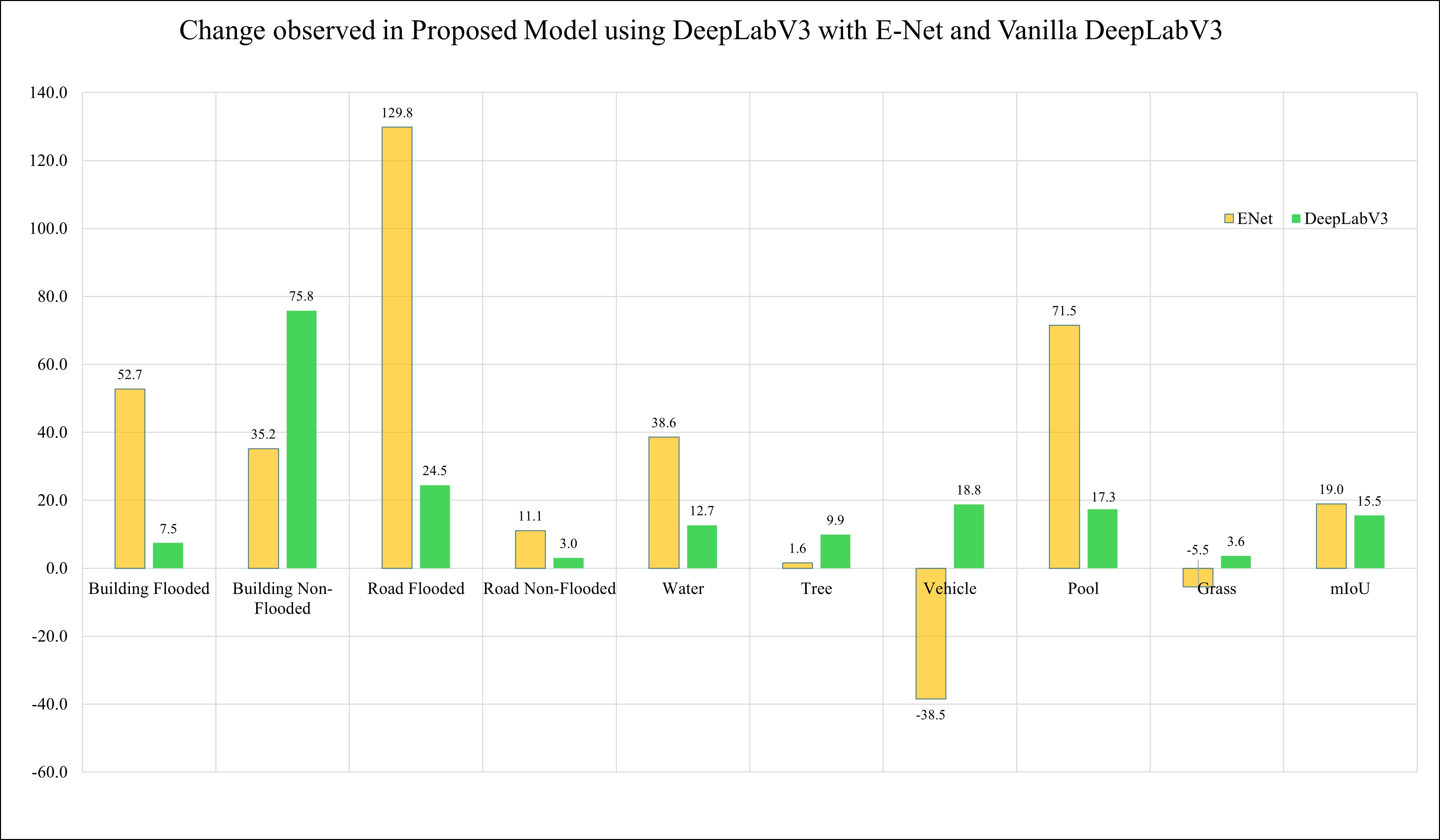}
        \caption{Relative changes observed between proposed model and DeepLabV3/E-Net}
        \label{fig:subfig_b}
    \end{subfigure}
    \caption{Relative changes observed in Proposed model using U-Net and DeepLabV3 with E-Net and U-Net/DeepLabV3}
    \label{fig:comp}
\end{figure}


The proposed method integrating DINOv2 features in DeepLabV3 achieved an improvement of 7.55 and 6.36 in terms of absolute mIoU performance when compared with E-Net and DeepLabV3, respectively. The significant improvement in the mIoU demonstrates the effectiveness of the DINOv2 features in accurately identifying the flooded and non-flooded regions in UAV regions.  The proposed method, in terms of class-wise IoU of 'Building Flooded', 'Building Non-Flooded, 'Road Flooded' and 'Road Non-Flooded' showed an improvement of 11.51, 14.58, 19.16 and 6.33, respectively, when compared with E-Net. Along with that, in terms of class-wise IoU specifically the 'Building Flooded', 'Building Non-Flooded, 'Road Flooded' and 'Road Non-Flooded' showed an increase in performance by 2.33, 24.15, 6.67 and 1.71, respectively, when compared to DeepLabV3. Relative changes in class-wise IoU and mIoU are also illustrated in Figure \ref{fig:subfig_b}.

Furthermore, the proposed DINOv2-U-Net achieved an improvement of 14.42 and 0.79 in terms of absolute mIoU performance when compared with E-Net and Vanilla U-Net, respectively. The proposed DINOv2-U-Net, in terms of class-wise IoU of 'Building Flooded', 'Building Non-Flooded, 'Road Flooded' and 'Road Non-Flooded' showed an improvement of 15.60, 28.51, 19.02 and 15.34, respectively, when compared with E-Net. Along with that, in terms of class-wise IoU specifically the 'Building Non-Flooded, 'Road Flooded' and 'Road Non-Flooded' showed an increase of 3.44, 1.23 and 0.33, respectively, when compared to Vanilla U-Net. Relative changes in class-wise IoU and mIoU are also illustrated in Figure \ref{fig:subfig_a}. These results demonstrate the effectiveness of the DINOv2 module in learning discriminant features, which helps in accurately identifying flooded and non-flooded regions from UAV aerial images. 

From this quantitative analysis, we observed that the proposed method using DeepLabV3 and DINOv2 achieved a significant performance increase in class-wise IoU for the first four classes compared to E-Net and DeepLabV3. Although the proposed method using U-Net did not significantly improve its performance in terms of class-wise IoU for the first four classes compared to Vanilla U-Net, it achieved significant improvements compared to E-Net. This improvement observed can be attributed to the fact that the DINOv2 model was trained on ADE20k \cite{ade} and Cityscapes \cite{cityscape} datasets, which had buildings, roads, routes, and vehicles as its classes, among many others. Although these image datasets and the model itself were trained on natural images, our analysis showed that the pre-trained model could be utilized for aerial images and can significantly improve performance. The results demonstrate that the features learned by the DINOv2 model using natural images can help in aerial image semantic segmentation and can help in learning better features during the training phase without the need for fine-tuning. 

\subsection{Qualitative Analysis}
This section compares the segmentation output of the proposed model with that of other models in terms of visual clarity, alignment with ground truth data, and contextual accuracy. The qualitative results are illustrated in Figure \ref{fig:qualitative_comparison}.

From Figure \ref{fig:qualitative_comparison}, we observed that DINOv2 with U-Net performed better than all the other models. Although all the models had difficulty segmenting the flooded roads, flooded buildings were correctly detected by the DINOv2 with U-Net model as observed in Figure \ref{fig:qualitative_comparison}c. Additionally, small objects such as vehicles and pools were accurately segmented, showcasing the model's ability to identify finer details. This level of performance highlights the robustness of the DINOv2 with U-Net in handling diverse scenarios, especially in comparison to other models, which struggled with either under-segmentation or over-segmentation in these challenging examples.
Both DINOv2 with U-Net (Figure \ref{fig:qualitative_comparison}c) and Vanilla U-Net (Figure \ref{fig:qualitative_comparison}d) produced smoother masks while Vanilla U-Net over-segmented in certain cases for trees and water in certain examples. 

Comparing the segmentation masks predicted by DeepLabV3 and DINOv2 with DeepLabV3, we observed that both these models struggled with separating overlapping regions, resulting in blurred or incomplete boundaries in certain cases. In addition to that, a lot of false negatives can also be observed. 

Overall, the qualitative analysis concurred with the quantitative results that a pretrained DINOv2 model with an encoder-decoder architecture for segmentation task produce better results. This also showed that the features learned by a pre-trained model on natural images are transferable to high-resolution UAV image semantic segmentation tasks. The results reinforce the importance of domain-agnostic DINOv2 features by training on a massive dataset and capturing patterns such as textures, shapes, and spatial relationships common across visual domains. These features, though derived from natural images, encode structural information (e.g., edges, boundaries, and color distributions) that aligns with aerial imagery components like roads, buildings, and vegetation.  


\subsection{Ablation Study on Loss Function}
\begin{table*}[htbp]
    \centering
    \begin{tabular}{|>{\centering\arraybackslash}p{2.0cm}|>{\centering\arraybackslash}p{1.5cm}|>{\centering\arraybackslash}p{1.2cm}|>{\centering\arraybackslash}p{1.2cm}|>{\centering\arraybackslash}p{1.0cm}|>{\centering\arraybackslash}p{1.2cm}|>{\centering\arraybackslash}p{0.8cm}|>{\centering\arraybackslash}p{1.2cm}|>{\centering\arraybackslash}p{0.8cm}|>{\centering\arraybackslash}p{1.0cm}|>{\centering\arraybackslash}p{0.8cm}|>{\centering\arraybackslash}p{1.0cm}|}
    
    \hline
        \textbf{Method} & \textbf{Loss} & \textbf{Building Flooded} & \textbf{Building Non-Flooded} & \textbf{Road Flooded} & \textbf{Road Non-Flooded} & \textbf{Water} & \textbf{Tree} & \textbf{Vehicle} & \textbf{Pool} & \textbf{Grass} & \textbf{mIoU} \\ \hline
        \textbf{DINOv2+U-Net} & Dice & 30.74 & 65.31 & 28.30 & 72.70 & 60.19 & 63.88 & 39.80 & 32.70 & 75.01 & 52.07 \\ \hline
        \textbf{DINOv2+U-Net} & Focal & 38.27 & 66.72 & 48.22 & 68.90 & 60.35 & 67.22 & 0.00 & 39.54 & 87.08 & 52.92 \\ \hline
        \textbf{DINOv2+U-Net} & Modified Focal & 41.84 & 69.23 & 35.97 & 68.10 & 66.22 & 68.50 & 20.36 & 30.70 & 79.03 & 53.33 \\ \hline
        \textbf{DINOv2+U-Net} & Dice + Modified Focal & 37.42 & 69.92 & 33.78 & 67.87 & 65.98 & 66.63 & 36.97 & 32.70 & 77.06 & 54.26 \\ \hline
        \textbf{DINOv2+} & Dice & 0.00 & 57.66 & 0.00 & 68.30 & 62.48 & 69.16 & 0.00 & 0.00 & 79.49 & 37.46 \\ 
        \textbf{DeepLabV3} & & & & & & & & & & &\\ \hline
        \textbf{DINOv2+} & Focal & 32.14 & 67.13 & 37.94 & 68.87 & 59.46 & 72.82 & 30.21 & 30.28 & 84.10 & 53.66 \\ 
        \textbf{DeepLabV3} & & & & & & & & & & &\\ \hline
        \textbf{DINOv2+} & Modified Focal & 35.12 & 64.70 & 28.84 & 63.92 & 60.86 & 66.96 & 27.21 & 29.13 & 74.56 & 50.15 \\
        \textbf{DeepLabV3} & & & & & & & & & & &\\ \hline
        \textbf{DINOv2+} & Dice + Modified Focal & 33.33 & 55.99 & 33.92 & 58.36 & 65.32 & 63.56 & 16.13 & 28.42 & 71.43 & 47.39 \\
        \textbf{DeepLabV3} & & & & & & & & & & &\\ \hline
    \end{tabular}
    \caption{Performance evaluation of the proposed approach with different loss functions (mentioned in the second column) in terms of per-class intersection over union (in \%). The mIoU value (in \%) is shown in the last column.}
    \label{ablation-study}
\end{table*}

As discussed earlier, the dataset suffers from class imbalance. Therefore, we conducted multiple experiments on the dataset to explore the effect of different loss functions on our two proposed approaches. Specifically, the performance with dice loss, focal loss, modified focal loss, and combined dice-modified focal loss was studied (Table \ref{ablation-study}).

In the case of DINOv2-U-Net, the evaluation of different loss functions highlights that the focal and modified focal loss consistently achieves higher IoU values compared to dice loss for Building Flooded, Building Non-Flooded, and Road Flooded classes, suggesting the effectiveness of these loss functions in segmenting classes with higher class imbalance, especially after adjusting the alpha values. Notably, a significantly higher IoU is obtained for the Building flooded and Building non-flooded classes for modified focal loss. However, focal loss struggles with the Vehicle class, obtaining an IoU of 0.00, which indicates its difficulty interpreting masks that are small in size. This issue is mitigated by modified focal loss, which improves the IoU to 20.36, and further addressed by dice loss, which achieves 39.80. Consequently, a combination of dice and modified focal loss presents a higher mIoU compared to all other loss functions.

Similarly, results for DINOv2-DeepLabV3 indicate that dice loss yields poor results for Building Flooded, Road Flooded, Vehicle, and Pool classes with an IoU of 0.00. Focal loss improves the performance across most classes and addresses the limitations of dice loss, notably attaining 32.14, 37.94, 30.21, and 30.28 respectively. Interestingly, modified focal loss also shows a significant improvement over dice loss, especially in building flooded and building non-flooded classes. However, a drop in IoU is observed for modified focal loss compared to focal loss in most classes. This drop could be due to architecture sensitivity to class imbalance and possibly due to the high-level semantic features extracted by DeepLabV3, which are already spread spatially.

\section{Conclusion}
This work investigates the feasibility of using self-supervised features learned from natural images to segment post-disaster aerial images. Specifically, we integrate self-supervised features from the pretrained DINOv2 model with two encoder-decoder-based segmentation approaches to identify flooded and non-flooded regions in UAV images. Our proposed method also introduces a class frequency-based custom focal loss to ensure that rarer classes, such as flooded regions, receive higher weights. This counteracts their under-representation during training.

We compare our approach with existing methods on the FloodNet dataset. The proposed architecture achieves a mean Intersection over Union (mIoU) score of 54.26 using the U-Net architecture and 47.39 using the DeepLabV3 architecture. While the DINOv2-DeepLabV3 model shows a significant improvement of 6.36

Our results validate the assumption that DINOv2’s self-supervised training on natural images produces transferable visual features that simplify aerial segmentation workflows. Additionally, utilizing self-supervised features from the pretrained DINOv2 model reduces the dependence on large, manually annotated datasets. This acceleration facilitates accurate and efficient processing of high-altitude imagery for real-world applications.
 
In the future, this work can be expanded to address the limitations found in class-wise Intersection over Union (IoU) scores for specific categories, such as pools, vehicles, and flooded roads. Additionally, various encoder-decoder architectures can be explored to assess their impact on performance. It would also be beneficial to investigate alternative learning strategies, like few-shot learning, to minimize reliance on manually annotated images. Furthermore, employing different loss functions within an ensemble approach could help reduce any existing discrepancies.


\section{Data Availability}
The dataset used in this study is available at \href{https://github.com/BinaLab/FloodNet-Supervised_v1.0}{https://github.com/BinaLab/FloodNet-Supervised\_v1.0}.

\section{Author Contribution Statement}

D.D. contributed towards Methodology, Implementation, Formal Analysis, Writing Original Draft, and Visualization. U.V. contributed towards Conceptualization, Methodology, Formal Analysis, Writing - Review \& Editing, and Supervision. All authors reviewed the manuscript.

    \small
    \bibliography{ref}

\end{document}